\documentclass[10pt,twocolumn,letterpaper]{article}

\usepackage[pagenumbers]{cvpr} 

\usepackage{algorithm}
\usepackage{algorithmic}
\usepackage{amsthm}
\newtheorem{condition}{Condition}
 \usepackage{amsmath}
 \usepackage{amssymb}
 \usepackage{tabularx}
 \usepackage{booktabs}
\usepackage{multirow}
\usepackage{graphicx}
\usepackage{colortbl}
\usepackage{array}
\usepackage{diagbox}

\usepackage{pifont}
\usepackage[table]{xcolor}
\newcommand{\yes}{\ding{51}}
\newcommand{\no}{\ding{55}}

\definecolor{cvprblue}{rgb}{0.21,0.49,0.74}
\usepackage[pagebackref,breaklinks,colorlinks,allcolors=cvprblue]{hyperref}

\title{CorrMem: Visually Grounded Correction Memory for Guideline-Consistent Segmentation}

\author{Vanshika Vats, Ashwani Rathee, James Davis\\
University of California, Santa Cruz, CA, USA\\
{\tt\small \{vvats,arathee1,davisje\}@ucsc.edu}
}

\begin{document}
\maketitle
\begin{abstract}
Guideline-consistent segmentation requires following detailed textual labeling policies, such as \textit{“include a pedestrian’s backpack in the mask, but exclude their wheeled suitcase.”} A multi-agent segmentation system that corrects such a guideline-specific mistake on one image should be able to avoid it on another. Yet existing refinement systems discard corrective feedback after each image, repeatedly spending computation to fix the same errors. We show that this feedback can become reusable guidance: local visual evidence can retrieve past corrections that improve predictions before refinement begins. We introduce CorrMem, a visually grounded correction memory that distills successful correction episodes into directive natural-language insights and anchors them to the image regions involved using patch-level visual features. When similar local content appears in a new image, dense patch matching retrieves the corresponding insights to guide the segmenting agent’s first prediction. This shifts the system from correcting recurring errors to preventing them. Even before any refinement, CorrMem achieves mDice gains of 1.5$\times$ and 2.1$\times$ over the baseline’s entire refinement loop on Waymo and Cityscapes, respectively. After refinement too, CorrMem outperforms the baseline across all reported metrics while using 2.6$\times$ and 1.9$\times$ fewer refinement steps. This shows that for agents making dense segmentation decisions, drawing on past experience grounded in local visual evidence can reduce recurring errors and the need for repeated refinement.
\end{abstract}

\section{Introduction}
Semantic segmentation systems are typically evaluated by whether they assign the correct semantic class to each pixel, whether through task-specific training \cite{ronneberger2015u, chen2018encoder} or, more recently, vision-language models (VLMs) prompted with text \cite{liang2023open, ren2024grounded}. In many real-world settings, however, correct segmentation depends on complex, task-specific labeling guidelines rather than visual category recognition alone. For example, a person pushing a bicycle may be labeled as a pedestrian, whereas one riding it may need to be excluded \cite{waymo}; an umbrella carried by a pedestrian may also need to be included in the same mask \cite{cityscapes}. Text-prompted segmentation systems excel on short prompts \cite{lisa, sun2025visual, ren2024grounded} but deteriorate under such long, intricate rules. Even capable models can therefore produce systematic guideline-specific errors in a single forward pass. Because the same visual patterns recur across images, the resulting errors do too. Crucially, a single-pass system has no mechanism to detect, let alone correct, its own mistakes.

Recent reasoning-driven segmentation methods increasingly move beyond single-pass prediction by decomposing inference into multiple stages, such as candidate generation and selection, interleaved reasoning and tool use, and explicit evaluation followed by prompt or mask refinement
\cite{zhu2025segagent,vats2026guideline,gao2026reason,he2025rsagent}.
These inference-time mechanisms can improve visual grounding and output reliability by allowing intermediate predictions to be inspected, compared, or revised. They often require multiple VLM or Multimodal LLM evaluations and tool invocations for each input sample. However, any corrective feedback they produce is generally confined to the current inference episode: once an image has been processed, its critiques and successful correction strategies are not retained for subsequent inputs. As a result, similar mistakes recur across images, forcing the system to incur additional refinement calls to rediscover a correction it could have reused on any later image with the same visual trigger given a fixed set of guidelines.

\begin{figure*}[ht]
  \centering
  \includegraphics[width=\linewidth]{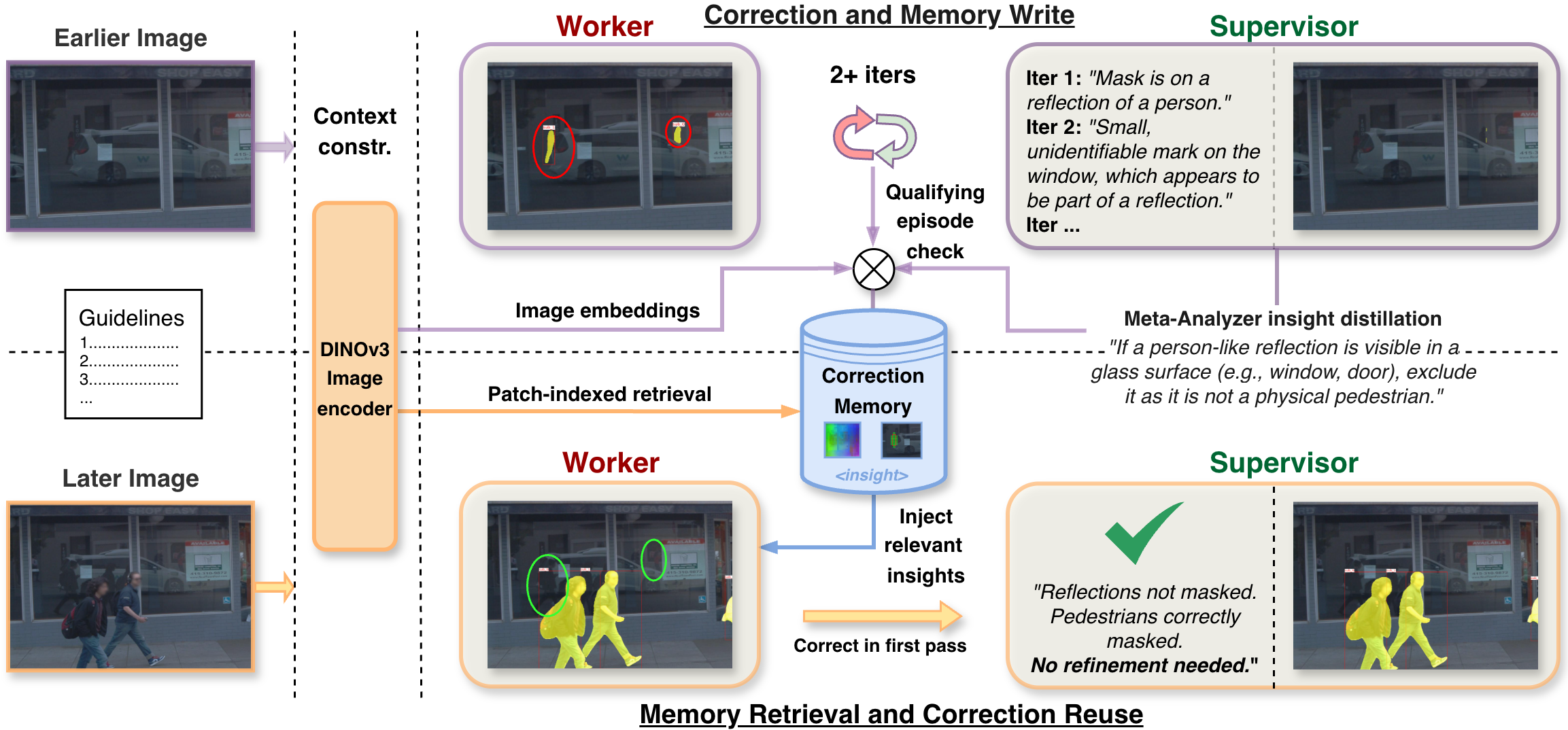}
  \caption{A simplified overview of CorrMem. The upper and lower panels illustrate two operations within the same online system: successful corrections from earlier images are distilled and written to a memory, while later images retrieve relevant insights via patch-level visual anchors before the initial prediction. This reuse prevents recurring guideline violations and reduces repeated refinement.}
  \label{fig:overall_pipeline}
\end{figure*}

A lesson like \textit{`reflections in glass storefronts are not physical pedestrians'} is conditional on a local visual cue and should be applied only when that pattern reappears (\cref{fig:overall_pipeline}). Existing methods apply past learnings using task text or whole-image similarity \cite{bo2025agentic, expel, suzgun-etal-2026-dynamic}, neither of which suits this setting: the complex guidelines are identical for every image, and a street scene's global appearance tells little about the reflection or umbrella that caused the error. Parametric fine-tuning is also a poor fit here because it requires curated data, risks interference with existing capabilities, and cannot readily incorporate corrections online. What this correction system needs is not better weights, but a memory designed around visual correction evidence.

To address this, we introduce \textit{CorrMem}, a patch-indexed correction memory that converts successful correction episodes into reusable, visually grounded insights. When a segmenter-critiquer refinement episode successfully finds and corrects an error, a meta-analyzer distills the episode into directive natural-language insights and anchors them to the relevant image regions using patch-level visual concept vectors. On subsequent images, this correction memory compares the stored concepts against dense patch embeddings, retrieves relevant insights, and gives them to the segmenting agent \textit{before its first prediction}. 

This shifts the system from repeatedly correcting mistakes to actively avoiding them in the first place, improving segmentation quality before any refinement occurs. Cleaner initial predictions also leave less work for the critiquer agent, reducing refinement iterations by 2.6$\times$ and 1.9$\times$ while improving final performance on Waymo \cite{waymo} and Cityscapes \cite{cityscapes} datasets, respectively. The memory is built online from an empty bank using the system's own correction episodes, without ground-truth masks, human corrective feedback, or gradient updates.

Thus, our main contribution is a visually grounded correction memory that connects natural-language lessons from past segmentation corrections to patch-level visual cues that indicate when those lessons should be reused. The guidelines specify the desired labeling behavior, while correction episodes reveal where the system struggles to follow them. CorrMem distills these episodes into directive insights and associates them with the visual regions involved in the correction. By retrieving these insights through dense patch matching before the initial prediction, CorrMem enables cross-image correction transfer, preventing recurring errors and reducing repeated refinement.
\section{Related Works}
\subsection{Multi-Agent Refinement in Segmentation}
Promptable segmentation conditions mask prediction on user-provided cues. These may be spatial prompts, such as points or boxes \cite{kirillov2023segany}, or language prompts, such as referring expressions and instructions used by unified vision-language segmentation models \cite{lisa, read}. Although effective with concise text prompts, these systems often struggle with long, intricate labeling policies that require multiple task-specific decisions. Even long-context VLMs such as Gemini \cite{comanici2025gemini, vats2026guideline} may fail to verify structured guideline compliance reliably in a single pass. A growing line of work therefore improves reliability through multi-stage generate-evaluate-revise inference. Self-Refine \cite{self_refine} and Reflexion \cite{reflexion} established this concept in language tasks, while some systems coordinate language and specialist models for multi-step tasks \cite{wu2023visual, shen2023hugginggpt}. In vision, prior works explore iterative localization \cite{Su2024ScanFormer, Sun2021IterativeShrinking}, self-correction and visual reasoning \cite{Liao2025SelfCorrection, sun2025visual}, and segmentation refinement \cite{vats2026guideline, he2025rsagent}. These methods improve predictions through additional inference stages, but operate within the current input and feedback is not retained for reuse across subsequent samples.

\subsection{Memory and Visual Retrieval}
Non-parametric memory offers an alternative to fine-tuning to retain and reuse experience. ExpeL \cite{expel} and Dynamic Cheatsheet \cite{suzgun-etal-2026-dynamic} accumulate insights for text-based reasoning, while other agents store reusable skills, workflows, or reasoning templates \cite{wang2024voyager, wang2025agent, yang2024buffer}. Existing multimodal memories typically rely on instruction similarity, task-observation pairs, or whole-image similarity \cite{sarch2024vlm, wang2024jarvis, bo2025agentic}. However, in guideline-consistent segmentation, the prompt and guidelines remain fixed across images, and the errors and their corrections depend on local visual cues. For example, for pedestrian segmentation, the applicable correction may depend on localized cues such as a reflection in a glass storefront, an umbrella carried by a person, or a human figure depicted on a poster. Relatedly, in-context segmentation transfers masks through patch-level correspondences either by prompting SAM \cite{liu2024matcher, zhang2024personalize}, or directly from frozen DINOv3 features \cite{cuttano2026insid3}. These methods require annotated reference masks and cannot encode policy-level corrections, such as \textit{`a person-shaped reflection must be excluded'}. 

CorrMem instead links natural-language corrections to patch-level visual features from regions where the system made mistakes. Annotated references cannot anticipate these regions: guidelines specify the rules, but the system’s own failures reveal where it struggles to follow them. \mbox{CorrMem} builds this memory online from its own critiques and retrieves relevant corrections when similar local trigger appears in a new image.

\section{Method}
\label{sec:method}

\subsection{Overview}
Guideline-consistent segmentation aims to produce a mask that is both visually accurate and compliant with the specified annotation guideline set, $\mathcal{G}$. The guidelines may specify long and fine-grained inclusion and exclusion rules, such as including objects carried by a pedestrian in the mask, while excluding reflections, mannequins, or people inside vehicles (complete guidelines in Supp. \cref{app:seg_guidelines}). We formulate the guideline-consistent segmentation within a Worker-Supervisor refinement framework \cite{vats2026guideline}, where the Worker VLM generates an initial segmentation, while the Supervisor VLM evaluates its compliance with $\mathcal{G}$ and provides feedback on missing instances, false positives, and boundary errors. An iteration controller uses the Supervisor’s feedback issue count to select whether to continue with the refinement loop. While this process corrects errors within an image, its feedback does not persist across next samples. CorrMem transforms this corrective feedback experience into persistent, visually grounded memory that can be reused across images.

We introduce a patch-indexed correction memory that lets this stateless segmentation system use its own correction insights and proactively apply that knowledge on new images before any mistake is made (Fig. \ref{fig:overall_pipeline}). Each memory starts empty. On the \textit{write} path, each qualifying episode is stored in the memory as a comprehensive instruction insight distilled by the meta-analyzer, paired with the specific visual anchors that supported it. On the \textit{retrieval} path, the new incoming image retrieves the relevant insights by dense patch-level appearance. These insights are finally inserted into the Worker's prompt as a high-priority past experience, helping it avoid known failures on the first pass itself. Ground truth is never used to build or query the memory, and the VLM and segmentation weights stay frozen.

We define an `episode' as a complete refinement cycle (one or more iterations) involving a worker producing the masks and a supervisor critiquing its work. To ensure a fair comparison, we follow prior work \cite{vats2026guideline} and quantify the supervisor's critique at iteration $t$ by issue count
\begin{equation}
    I_{t} = I_{miss} + I_{false} + 0.1I_{ref}
\end{equation}
where $I_{miss}$, $I_{false}$, and $I_{ref}$ are the number of missing objects, false positives, and segmentation boundary refinements flagged, respectively. Refinements are down-weighted as they indicate mask-quality adjustments rather than detection errors. These issue counts are further used by the qualifying-episode check to determine whether an episode qualifies for processing by the meta-analyzer.

\subsection{Memory Write}
\label{sec:mem_write}

\paragraph{Qualifying-Episode Check.}
\label{sec:learning_event}
Not every worker-supervisor correction cycle provides a transferable lesson. Some runs require no refinement, while others involve only minor mask adjustments or even degrade despite feedback. Such cases do not reveal a reliable correction pattern that can generalize beyond the current image and would only bloat our memory unnecessarily. We therefore admit an episode to the write path only if it is a \textit{qualifying episode}, computed from its issue-count trajectory ($I_{0}, I_{1}, \ldots, I_{T}$) over refinement passes. 
An episode is qualified to provide a lesson iff:
\begin{condition}
The total issue reduction satisfies $I_{0} - I_{T} \geq 1$, i.e., the score decreased by at least one unit, filtering out episodes driven only by minor boundary fluctuations.
\end{condition}
\begin{condition}
No single refinement step regressed by more than $\delta$, i.e. $ I_{t} - I_{t-1} \leq \delta, \forall t \geq 1$.
\end{condition}
The first condition ensures the episode contains a genuine, non-trivial correction; the second rejects unstable trajectories whose net improvement may be incidental. A $\delta = 0.5$ tolerates noisy $I_{ref}$. Episodes with a clean initial critique skip refinement entirely and never reach this gate. Only episodes that pass this qualifying check are handed to the meta-analyzer; all others are discarded.


\paragraph{Meta-Analyzer.}
\label{sec:meta_analyzer}
For each qualifying episode, the meta-analyzer receives the original image and the supervisor’s missing-object and false-positive correction items, each comprising a localized bounding box and its rationale. This provides it with a holistic view of the scene and how the corrections unfolded. In a single batched VLM call, the meta-analyzer distills these items into one or more insights. Each insight is a
directive rule that names a visual cue and the labeling action it requires. 

A valid insight must satisfy two constraints: its cited trigger must be visually verifiable in the image, and its corrective rule must be non-trivial and generalizable. The meta-analyzer then associates each insight with the subset of Supervisor-provided boxes that supports it; it does not generate new boxes. This produces tuples of the form \textit{(insight text, support regions)}, grounding each insight in the visual evidence that motivated it. We use these support regions to attach visual grounding to each insight.

\begin{figure}[t]
  \centering
  \includegraphics[width=\linewidth]{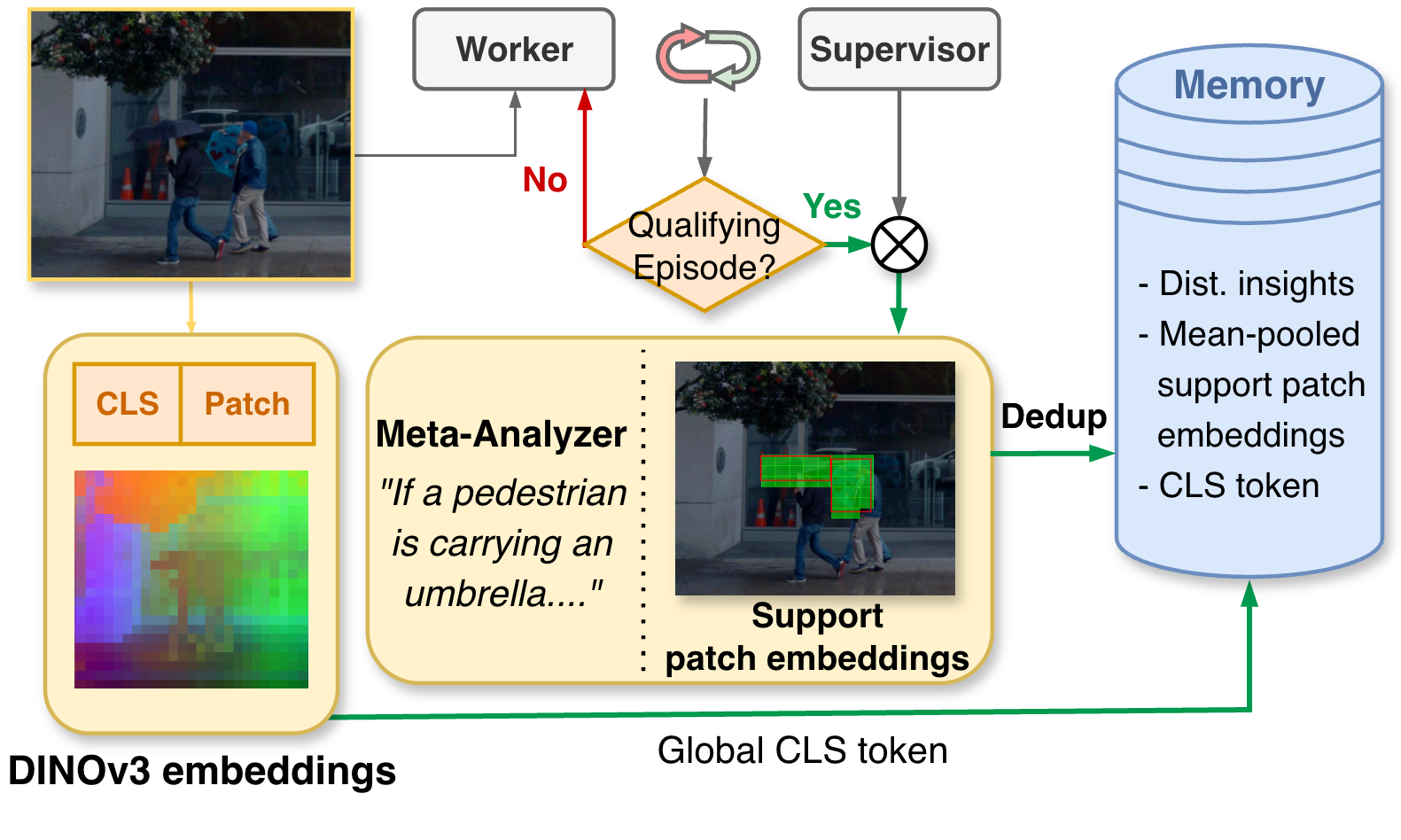}
  \caption{Memory writing. Meta-analyzer distills each qualifying episode into natural-language insights which are visually anchored with patch tokens, mean-pooled over the support regions (\textit{umbrellas} in the figure) to form a concept vector and written to the memory. The CLS token is used only to keep anchors diverse.}
  \label{fig:store_memory}
\end{figure}

\paragraph{Support-Region Patch Embeddings.}
\label{sec:patches}
For each insight, we capture the \textit{local} appearance of the error regions that triggered refinement feedback using patch embeddings from DINOv3 \cite{dinov3} image encoder. Each image encoded at $512\times512$ passes through this, producing a $32\times32$ grid of L2-normalized patch tokens. Each patch token represents a local visual embedding of its corresponding image region.

An insight's \textit{support regions} are the subset of supervisor-provided boxes associated with it by the meta-analyzer. Its \textit{support patches} are the DINOv3 patch tokens whose grid cells overlap with those boxes by at least 30\% (\cref{fig:store_memory}). The 30\% overlap threshold is intentional: allowing patches to fall partly outside the boxes helps capture not only the object trigger, but also its surrounding context.

A single insight is often supported by several regions with different appearances. For example, for an insight \textit{`if a person is carrying a bag in their hand, do...'} there could be several trigger `bag' support regions in an image. However, retrieval needs one stable representation of the underlying trigger concept. We therefore mean-pool the selected support patches across support regions and L2-normalize the result to form the trigger's \textit{concept vector}. Because these patches come from the regions the system actually mis-segmented rather than from random sampling, the concept vector reflects a genuine past failure.

\begin{figure}[t]
  \centering
  \includegraphics[width=\linewidth]{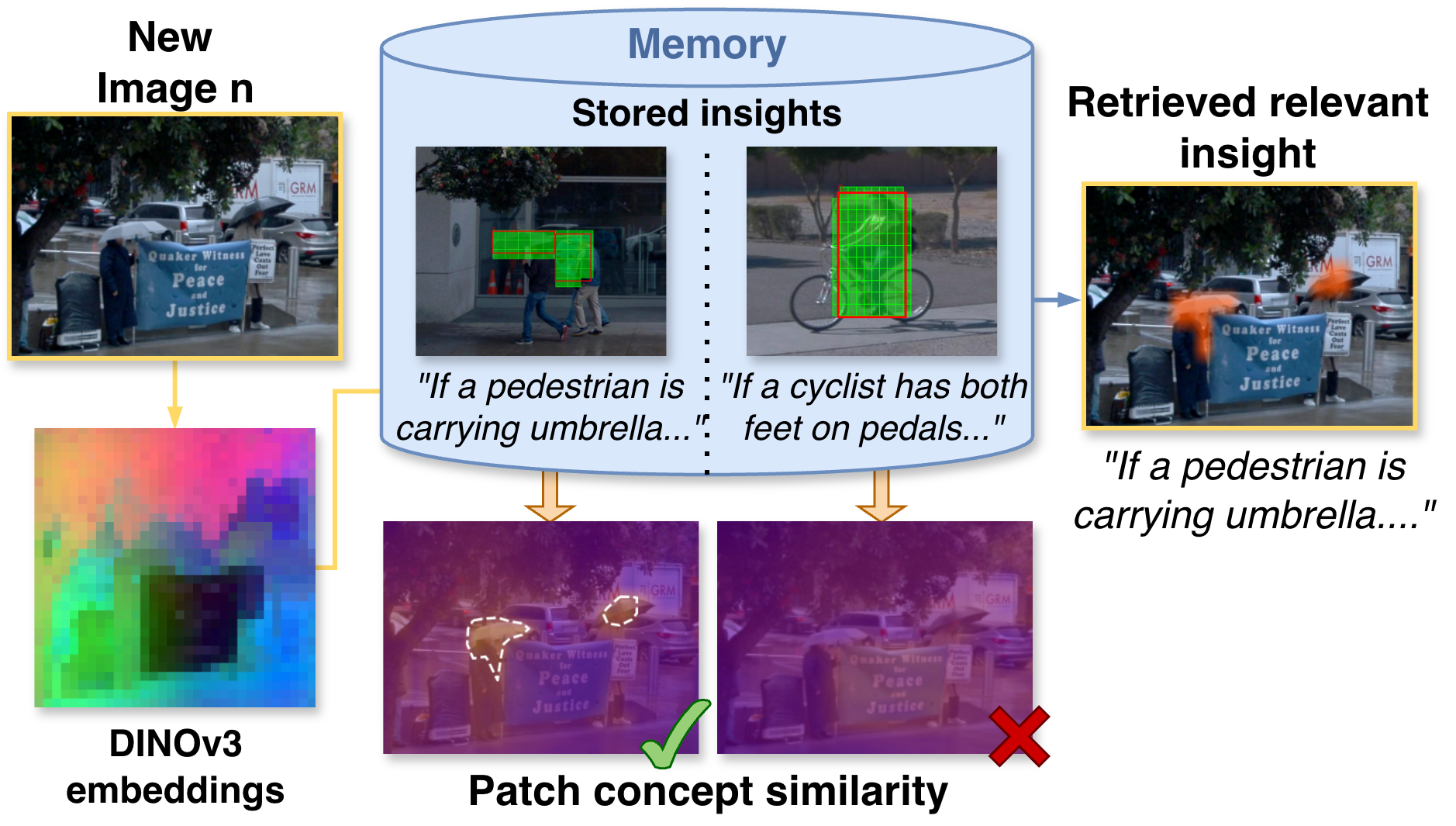}
  \caption{Memory retrieval. Patch tokens of a new image are compared with the concept vectors of all stored anchors. Similarity is computed per patch, so an insight fires when its concept matches a local region rather than the whole scene. The top-scoring insights are injected into the Worker's prompt before its initial prediction.}
  \label{fig:retrieval}
\end{figure}

\paragraph{Insight deduplication.}
Stored insights help the system avoid repeating mistakes it has already experienced. However, a new scene can still sometimes trigger a similar mistake, leading the meta-analyzer to produce a lesson that closely matches one already in memory. Thus, new insights are first deduplicated before storing. An insight's text embedding is compared against those of stored insights by cosine similarity. If the closest match exceeds a high threshold of 0.90, we treat the insight as a restatement of an existing record and add its anchor as a candidate to that record. If not, we create a new record.

\paragraph{Memory Storage.}
\label{sec:storage}
The memory bank stores a set of unique insights, each grounded by one or more \textit{visual anchors}. An anchor is a record of a single valid episode containing a concept vector for local information and a DINOv3 CLS token for global scene information (Fig. \ref{fig:store_memory}). During retrieval, we use only the patch-level concept vector. The CLS embedding is used only to maintain visual diversity among an insight's anchors, as described below. Since one pooled concept vector cannot cover every possible appearance of a trigger (e.g. handbags, backpacks etc. for a trigger `bag'), each insight keeps up to $K_a=5$ visually diverse anchors. This lets memory represent the same concept through several examples rather than just one pooled example.

\paragraph{Anchor admission.} 
We want the anchors of an insight to cover different visual forms of the same underlying trigger, rather than repeatedly storing near-identical evidence. For example, a `carried bag' insight is more useful if its anchors include bags under different lighting, weather, viewpoints, and surrounding scenes, instead of five nearly identical sunny street examples. A new anchor is added only if its scene is visually different from the anchors already stored, measured using the global CLS embedding cosine similarity match. If it is too similar, we reject its admission. If it passes and there is space in the record $K_a$, we include it. If the record is already full, it replaces the stored anchor most similar to it, keeping the five anchors from clustering around near-duplicate scenes.

\subsection{Memory Retrieval}
When a new sample arrives, retrieval happens once before the Worker's initial pass (\cref{fig:retrieval}). The query image is encoded by the same DINOv3 backbone into a grid of L2-normalized patch tokens, stacked as $Q \in \mathbb{R}^{P^2 \times d}$, where $P=32$, and $d$ is the patch-embedding dimension. Each stored anchor has a concept vector obtained by mean-pooling the patch tokens inside its support regions and L2-normalizing the result. Let $C \in \mathbb{R}^{M \times d}$ stack the concept vectors of all $M$ anchors in the bank. We compute the patch-concept similarity matrix
\begin{equation}
S = QC^\top \in \mathbb{R}^{P^2 \times M},
\end{equation}
where $S_{ij}$ is the cosine similarity between query patch $i$ and anchor concept vector $j$. Each anchor is scored by its strongest match over the query patches
\begin{equation}
a_j = \max_{1 \le i \le P^2} S_{ij}.
\end{equation}
Thus, an anchor fires when a local region of the query image resembles the visual concept encoded by that anchor. Each insight $ins$ then takes the score of its best-firing anchor
\begin{equation}
r_{\mathrm{ins}} = \max_{j \in \mathcal{J}_{\mathrm{ins}}} a_j,
\end{equation}
where $\mathcal{J}_{\mathrm{ins}}$ indexes the anchors associated with $ins$. We rank insights by $r_{\mathrm{ins}}$ and retrieve the top $k_r$ insights whose scores satisfy $r_{\mathrm{ins}} \ge \tau_{\mathrm{ret}}$. Retrieving a small, fixed set keeps the Worker's prompt bounded as the memory grows online, while ensuring that each injected insight is supported by visual evidence in the current image.\
\\

\begin{figure*}[ht]
  \centering
\includegraphics[width=\linewidth]{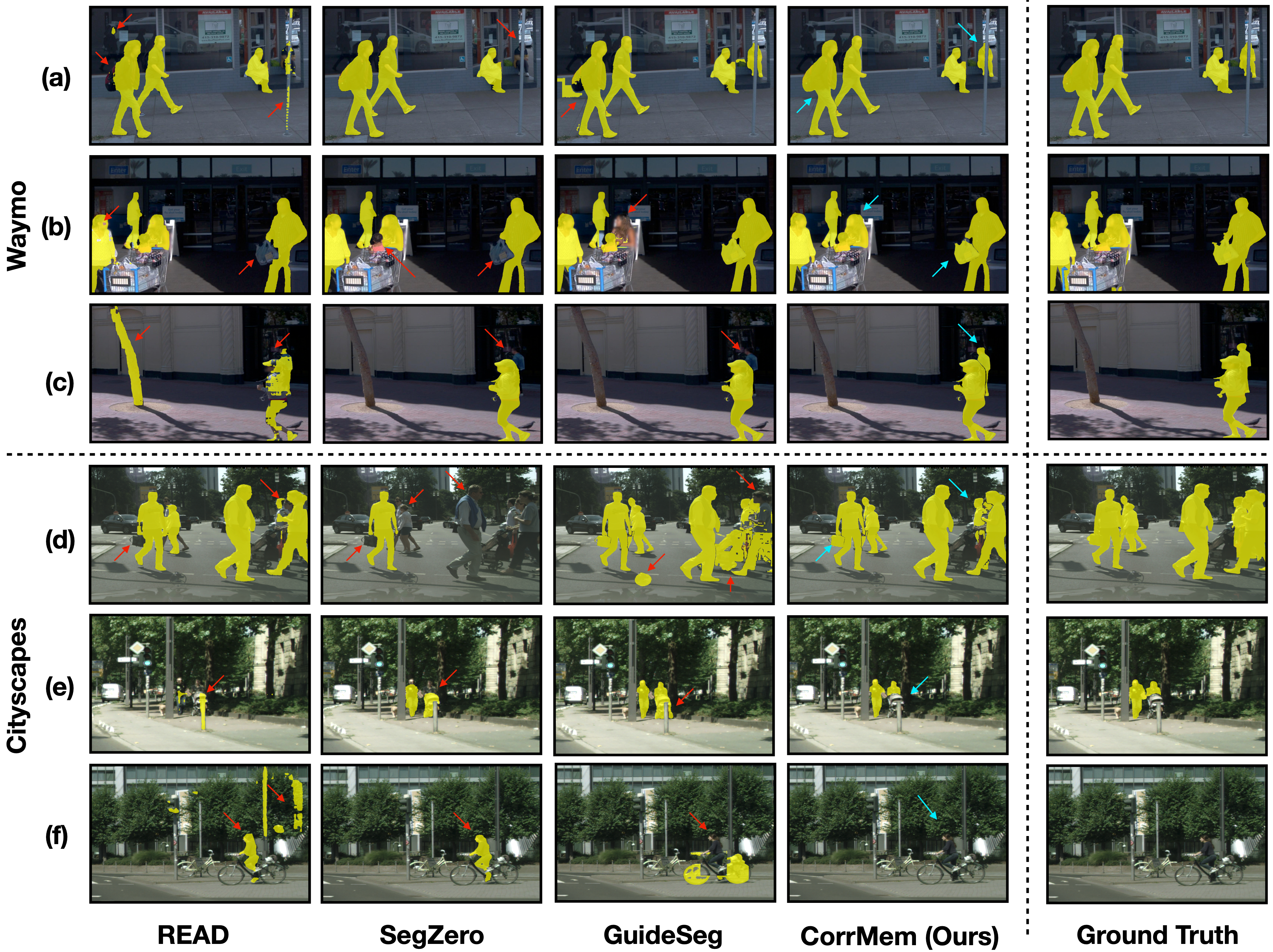}
  \caption{Qualitative results on both the Waymo and Cityscapes datasets. Yellow denotes the predicted segmentation masks, with ground truth shown in the final column. Red arrows highlight representative segmentation errors in the compared methods, while blue arrows indicate challenging regions correctly handled by CorrMem.}
  \label{fig:qualitative}
\end{figure*}

\noindent The retrieved insight texts are injected once into the Worker's initial prompt as a high-priority `past experience' block, with the instruction to apply each insight only when its described visual cue is actually present in the image. Retrieval does not disable the write path. Even when an insight is retrieved for the current image, the resulting episode is still evaluated by the memory-admission criterion, and any qualifying correction is added/merged with the memory.

\section{Experiments}

\subsection{Datasets}
We evaluate our method on \textit{Waymo guideline-consistent} and \textit{Cityscapes} datasets. The Waymo guideline-consistent dataset by Vats et al. \cite{vats2026guideline} contains 101 carefully curated samples whose annotations strictly follow the defined labeling rules. Since the study's goal is to enforce guideline-consistency, this dataset filters out the noisy ground-truth annotations that deviate from Waymo’s own labeling protocol. To evaluate robustness beyond the curated benchmark, we also report results on the \textit{Cityscapes} \cite{cityscapes} 500 sample validation set. Similar to prior works, we focus on the \textit{Pedestrian} class (\textit{Person} class in Cityscapes), as it presents a more challenging segmentation problem due to subtle object boundaries, complex guidelines, and frequent annotation ambiguities. In contrast, larger classes such as \textit{Car} tend to inflate aggregate metrics while hiding fine-grained segmentation errors. Additional experiments with the \mbox{\textit{Cyclist}} class and the complete guideline sets are provided in Supp. \cref{app:cyclist}, \cref{app:seg_guidelines}.

\subsection{Implementation}
For an input image $Img_i$ and guidelines $\mathcal{G}$, we encode each image using a DINOv3 ViT-H+/16 backbone to extract image embeddings. We use the stable gemini-2.5-flash as the same base model as the stateless benchmark \cite{vats2026guideline}. The Worker VLM is run with temperature $T=0.5$ to allow flexibility for object localization, while the Supervisor VLM is run more deterministically with $T=0.3$. SAM-2.1-Hiera-Large is used to generate the segmentation masks.

Conditions 1 and 2 (\cref{sec:method}) determine whether a correction episode is qualifying. If the event passes this check, the Supervisor's feedback is distilled by a meta-analyzer VLM ($T=0.3$) into a compact insight and written to a memory bank implemented with FAISS \cite{johnson2019billion}. Memory insertion is controlled by a text-level deduplication via bge-base-en-v1.5 \cite{xiao2024c}. During retrieval, we set $\tau_{\mathrm{ret}}=0.6$ and return the top $k_r=3$ insights. The thresholds are selected through preliminary experiments on data separate from the evaluation sets and kept fixed across all datasets and settings.

The memory is initialized empty. For an image, retrieval uses only insights from previously processed images, and insights are written only after its final prediction. Ground truth is used solely for evaluation.

Following the initial Worker prediction, a complete refinement iteration consists of a Supervisor-Worker exchange: the Supervisor critiques the current mask, and the Worker applies the feedback to produce a revised mask. If the Supervisor identifies no issues, the event terminates with zero refinement iterations. We set the maximum number of iterations to 4 \cite{vats2026guideline}. All local inference is performed on 3$\times$RTX 3080 GPUs. We conduct three independent runs with random seeds and report means and standard deviations for principal metrics.

\subsection{Evaluation Metrics}
We use global IoU (gIoU) and mean Dice (mDice) as segmentation metrics \cite{kazemzadeh-etal-2014-referitgame, giou_ciou, lisa}. gIoU averages per-image IoU, giving each image equal weight, while mDice measures mask overlap and captures the trade-off between over/under-segmentation. We also report mean precision (mPr) and recall (mRec) to assess these errors separately, alongside cumulative IoU (cIoU), which is disproportionately more influenced by larger objects than gIoU.

\begin{table*}[t]
\centering
\caption{Segmentation results on the Waymo \textit{guideline-consistent} dataset and Cityscapes \textit{val} set. CorrMem outperforms prior methods on all metrics across both datasets. The highest numbers are bolded, and the second highest underlined.}
\begingroup
\footnotesize
\renewcommand{\arraystretch}{1.25}
\setlength{\tabcolsep}{3pt}

\newcolumntype{C}{>{\centering\arraybackslash}X}

\begin{tabularx}{\textwidth}{@{}p{0.14\textwidth}|>{\centering\arraybackslash}p{0.18\textwidth}|*{5}{C}|*{5}{C}@{}}
\toprule
\multirow{2}{*}{Method} &
\multirow{2}{*}{Backbone} &
\multicolumn{5}{c|}{Waymo \textit{guideline-consistent}} &
\multicolumn{5}{c}{Cityscapes \textit{val}} \\
\cmidrule(lr){3-7}\cmidrule(l){8-12}
& &
\multicolumn{1}{c}{gIoU} &
\multicolumn{1}{c}{cIoU} &
\multicolumn{1}{c}{mPr} &
\multicolumn{1}{c}{mRec} &
\multicolumn{1}{c|}{mDice} &
\multicolumn{1}{c}{gIoU} &
\multicolumn{1}{c}{cIoU} &
\multicolumn{1}{c}{mPr} &
\multicolumn{1}{c}{mRec} &
\multicolumn{1}{c}{mDice} \\
\midrule

LISA-7B \cite{lisa} & {\scriptsize LLaVA-7B + SAM-H} &
23.41 & 18.32 & 24.38 & 79.79 & 33.17 &
30.56 & 42.15 & 46.90 & 38.19 & 39.90 \\

LISA-13B \cite{lisa} & {\scriptsize LLaVA-Llama2-13B + SAM-H} &
20.16 & 14.21 & 22.41 & 62.41 & 28.30 &
26.75 & 36.11 & 39.65 & 36.65 & 35.42 \\

GroundedSAM \cite{ren2024grounded} & {\scriptsize Grounding DINO-T + SAM-H} &
20.43 & 19.36 & 29.35 & 28.11 & 25.63 &
21.31 & 05.53 & 33.31 & 40.04 & 28.32 \\

READ \cite{read} & {\scriptsize LLaVA-1.5-13B + SAM-H} &
43.94 & 45.55 & 51.74 & 69.75 & 59.41 &
24.04 & 36.60 & 35.26 & 35.13 & 32.44 \\

SAM3-Agent \cite{sam3} & {\scriptsize Qwen3-VL-8B + SAM3} &
61.67 & 61.74 & 82.27 & 64.00 & 70.58 &
\underline{51.76} & \underline{65.61} & 65.81 & \underline{63.41} & \underline{62.06} \\

Gemini-2.5 \cite{comanici2025gemini} & {\scriptsize Gemini-2.5-Flash+SAM2.1-L} &
69.02 & 74.24 & 80.91 & 79.89 & 78.75 &
35.62 & 46.87 & 47.94 & 54.62 & 46.29 \\

SegZero \cite{segzero} & {\scriptsize Qwen2.5-VL-7B + SAM2-L} &
71.96 & 73.72 & 86.34 & 78.45 & 80.88 &
41.52 & 43.47 & 60.75 & 51.36 & 52.78 \\

GuideSeg \cite{vats2026guideline} & {\scriptsize Gemini-2.5-Flash + SAM2.1-L} &
\underline{80.57} & \underline{86.70} & \underline{91.06} & \underline{84.78} & \underline{87.20} &
51.14 & 64.55 & \underline{66.43} & 61.07 & 61.72 \\

\midrule
\rowcolor[HTML]{EFEFEF}
\textbf{CorrMem (Ours)} & {\scriptsize Gemini-2.5-Flash + SAM2.1-L} &
\textbf{83.51} & \textbf{87.68} & \textbf{94.44} & \textbf{87.55} & \textbf{90.21} &
\textbf{55.71} & \textbf{70.40} & \textbf{72.89} & \textbf{63.44} & \textbf{66.35} \\
\bottomrule
\end{tabularx}

\endgroup
\label{tab:main-results-waymo-cityscapes}
\end{table*}
\begin{table}[ht]
\renewcommand{\arraystretch}{1.2}
\centering
\caption{Effect of utilizing memories on the Waymo dataset. By using previous correction insights, our method improves the initial segmentation quality and further surpasses the refinement baseline after refinement, while reducing the average refinement passes by more than half.}
\resizebox{\columnwidth}{!}{%
\begin{tabular}{@{}r|cc|cc@{}}
\toprule
\multirow{2}{*}{\diagbox[width=2.0cm,height=2.3\line,outerleftsep=1pt,outerrightsep=0pt,innerleftsep=0pt,innerrightsep=0pt]{Metrics}{Method}} &
\multicolumn{2}{c|}{Before Refinement} &
\multicolumn{2}{c}{After Refinement} \\
\cmidrule(l){2-3}\cmidrule(l){4-5}
& No memory & Ours & No memory & Ours \\
\midrule
gIoU $\uparrow$  & 73.87 & \textbf{78.85} & 80.57 & \textbf{83.51} \\
cIoU $\uparrow$ & 78.40 & \textbf{82.88} & 86.70 & \textbf{87.68} \\
mPr $\uparrow$  & 88.36 & \textbf{90.31} & 91.06 & \textbf{94.44} \\
mRec $\uparrow$ & 80.81 & \textbf{86.58} & 84.78 & \textbf{87.55} \\
mDice $\uparrow$ & 84.75 & \textbf{88.40} & 87.20 & \textbf{90.21} \\
\midrule
\rowcolor[HTML]{EFEFEF}
\multicolumn{1}{r|}{Avg \#iters $\downarrow$} &
- & - &
2.66 & \textbf{1.01} \footnotesize{\textit{($\downarrow2.6\times$)}}\\
\bottomrule
\end{tabular}%
}
\label{tab:num_iter}
\end{table}
\subsection{Results}
We compare our system against text prompt-conditioned segmentation methods with their code available, such as LISA \cite{lisa}, GroundedSAM \cite{ren2024grounded}, SAM3-Agent \cite{sam3}, READ \cite{read}, Gemini-2.5 \cite{comanici2025gemini}, SegZero \cite{segzero}, and GuideSeg \cite{vats2026guideline}, under the same guidelines. As demonstrated in \cref{tab:main-results-waymo-cityscapes}, on the \textit{Waymo} guideline-consistent benchmark, our method achieves 83.51($\pm$0.26) gIoU and 90.21($\pm$0.45) mDice, outperforming the strongest refinement baseline, GuideSeg with the same underlying VLM, by +2.94 and +3.01 points, respectively. The improvement is consistent across all Waymo metrics, with gains of +3.38 in precision and +2.77 recall. This shows that CorrMem improves both sides of the segmentation trade-off: it recovers more guideline-relevant objects while also reducing false positives. Similar trend holds on \textit{Cityscapes} val, where our method improves over its strongest baseline, SAM3-Agent by +3.95 gIoU, +4.79 cIoU, and +4.29 mDice. This setting is particularly challenging because many object instances are very small and distant, making the visual cues needed for precise segmentation weaker. Qualitative examples of both datasets are shown in \cref{fig:qualitative}.

\subsection{Discussion and Analysis}
\paragraph{{Effect of Reusable Memory.}} 
To examine the contribution of our cross-image memory, \cref{tab:num_iter} compares the segmentation pipeline with and without reusable insights, both before and after refinement. Before any refinement, our method already improves over no-memory baseline by +4.98 gIoU and +3.65 mDice. At this point no supervisor feedback has touched the current image, so this first-pass gain can only come from insights transferred from previously seen images. Importantly, the first-pass gIoU improvement is statistically reliable: on the most conservative run (seed 1311), the gIoU gain is +3.6 (paired-bootstrap 95\% CI [+1.3, +5.9]; Wilcoxon signed-rank $p<0.001$). Evaluating on a held-out set with a frozen memory bank, the first-pass still improves by +7.66 gIoU when using memory (Supp. \cref{app:held_out}). This shows that insights accumulated on one set of images can transfer to unseen scenes from the same dataset without further memory updates. After refinement, our method remains better while requiring substantially fewer passes, reducing the average number of refinement iterations by $2.6\times$. Thus, the system reaches higher final segmentation quality while requiring fewer refinement calls. Similar improvements are shown for the Cityscapes dataset in Supp.\cref{tab:num_iter_cityscapes} with $1.9\times$ fewer steps. The overall gain is also robust to the sample processing order \mbox{(Supp. \cref{tab:split-comparison-before-after})}. The memory remains economical (under 2 MB per benchmark; Supp. \ref{cost}): deduplication substantially compresses the insights produced by the meta-analyzer on both benchmarks, merging 47.6\% of 42 on Waymo and 68.3\% of 104 on Cityscapes. The compact memory entries retain an average of 1.5 and 2.0 visual anchors per insight, respectively.

This turns refinement from a purely reactive process into a memory-guided one. In a stateless refinement system, the Worker first makes a prediction, the Supervisor identifies the errors, and additional refinement steps are needed. Each new sample, therefore, starts from scratch, even when it contains mistakes the system has already corrected before. Our method makes this process proactive, letting the Worker avoid recurring guideline errors before they enter the refinement loop. Overall, these results show that reusable correction insights improve segmentation quality and efficiency together by preventing the system from rediscovering the same guideline-specific mistakes.

\paragraph{Patch-Level vs Global Retrieval.}
We conduct all further ablations on the Waymo guideline-consistent dataset. We first test whether global scene-level retrieval is sufficient by replacing the local patch-based concept vectors with DINOv3 CLS embeddings for both memory indexing and retrieval. As shown in \cref{tab:cls_patch_ablation}, using global CLS provides no consistent improvement in segmentation quality: it increases precision and Dice, while degrading gIoU, cIoU, and recall relative to the stateless baseline. In contrast, patch-based retrieval improves all segmentation metrics. This indicates that whole-image similarity captures broad scene appearance but often misses the localized visual cues that determine guideline relevance. Dense patch matching instead recalls insights based on their specific visual triggers, enabling more precise and cue-specific retrieval. Qualitative examples are provided in Supp. ~\cref{fig:cls_vs_patch}.


\begin{table}[t]
\footnotesize
\centering
\caption{Comparison of local and global visual retrieval. Patch-based retrieval consistently outperforms global CLS retrieval.}
\renewcommand{\arraystretch}{1.2}
\resizebox{\columnwidth}{!}{%
\begin{tabular}{@{}l|ccccc@{}}
\toprule
\multicolumn{1}{c|}{Memory} & 
gIoU$\uparrow$ & cIoU$\uparrow$ & mPr$\uparrow$ & mRec$\uparrow$ & mDice$\uparrow$ \\ 
\midrule
None        & 80.57 & 86.70 & 91.06 & 84.78 & 87.20 \\
Global CLS  & 80.22 & 84.06 & 93.33 & 84.51 & 87.99 \\
Patch-based & \textbf{83.51} & \textbf{87.68} & \textbf{94.44} & \textbf{87.55} & \textbf{90.21} \\
\bottomrule
\end{tabular}%
}
\label{tab:cls_patch_ablation}
\end{table}

\begin{table}[t]
\centering
\caption{Robustness to irrelevant memory insertion. Injecting the system with least-relevant memories performs similarly to no memory, showing that the system does not blindly apply unsupported insights. Most-relevant memories improve first-pass performance across all metrics.}
\footnotesize
\renewcommand{\arraystretch}{1.25}
\setlength{\tabcolsep}{5pt}

\newcolumntype{C}[1]{>{\centering\arraybackslash}m{#1}}

\begin{tabular}{@{}C{0.20\columnwidth}|C{0.15\columnwidth}|C{0.22\columnwidth}|C{0.30\columnwidth}@{}}
\toprule
\multirow{2}{*}{\diagbox[width=0.20\columnwidth,height=2.5\line,%
  outerleftsep=1pt,outerrightsep=0pt,innerleftsep=1pt,innerrightsep=0pt]%
  {Metric}{Method}} &
\multicolumn{3}{c}{Memory} \\
\cline{2-4}
\addlinespace[3pt]
&
None &
\mbox{Least-relevant} &
\mbox{Most-relevant} \\
\midrule
gIoU $\uparrow$  & 73.87 & 73.27 & \textbf{78.85} \textit{(+4.98)} \\
cIoU $\uparrow$  & 78.40 & 77.32 & \textbf{82.88} \textit{(+4.48)} \\
mPr $\uparrow$   & 88.36 & 87.70 & \textbf{90.31} \textit{(+1.95)} \\
mRec $\uparrow$  & 80.81 & 81.07 & \textbf{86.58} \textit{(+5.77)} \\
mDice $\uparrow$ & 84.75 & 84.25 & \textbf{88.40} \textit{(+3.65)} \\
\bottomrule
\end{tabular}
\label{tab:memory-relevance}
\end{table}

\begin{figure}[t]
  \centering
  \includegraphics[width=0.95\linewidth]{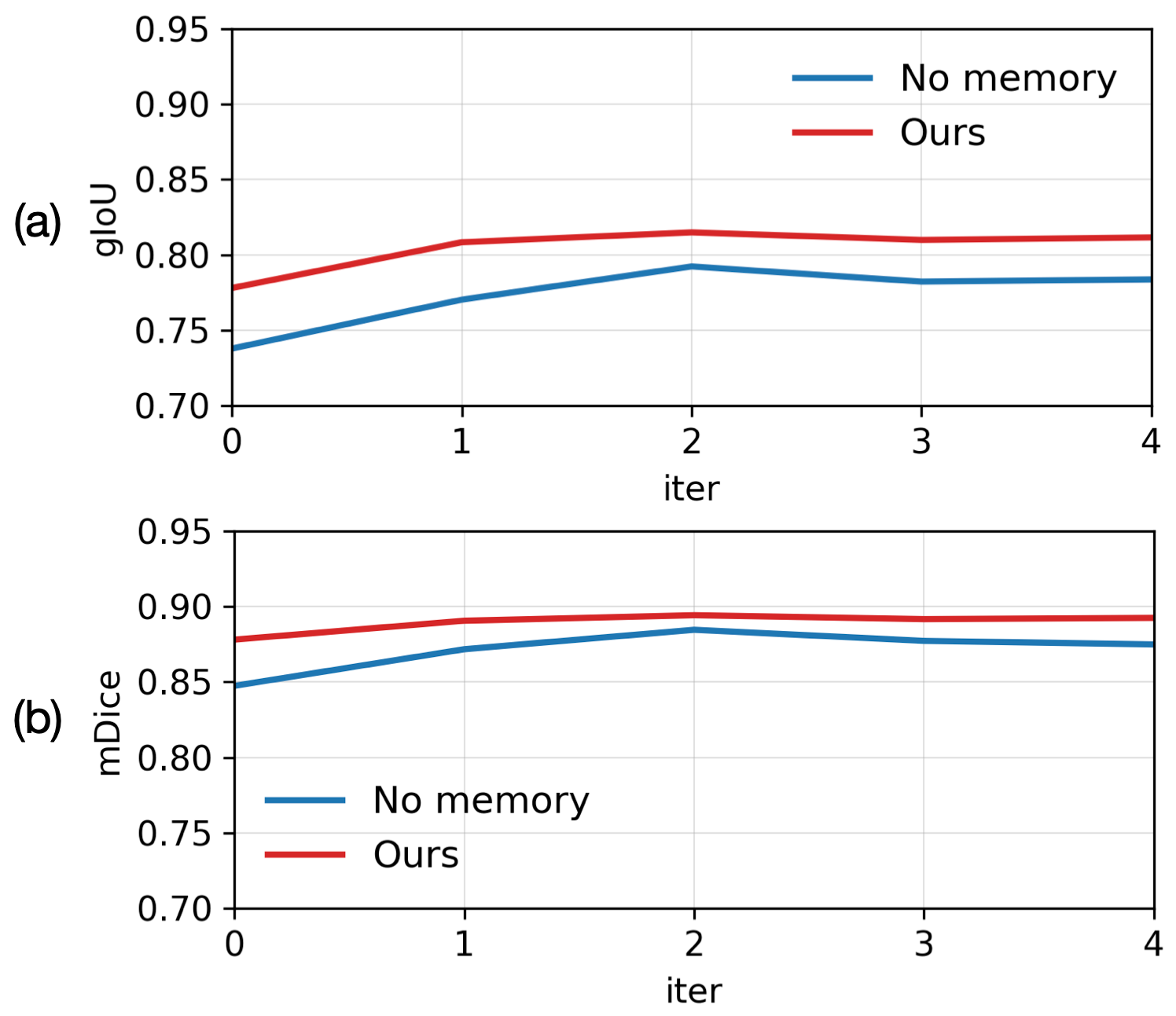}
  \caption{Performance across refinement iterations without early stopping. Our method produces stronger initial predictions and remains consistently above the stateless refinement baseline, showing that reusable memory improves both the first-pass output and the refined predictions.}
  \label{fig:all_iter_ablation}
\end{figure}

\paragraph{Memory Gains Persist Across Refinement Passes.}
To test whether the no-memory baseline catches up if given more refinement, we
remove the stopping criterion and run both systems up to the maximum refinement budget. \cref{fig:all_iter_ablation} shows that it does not. CorrMem starts from a stronger initial prediction and remains consistently above the baseline across all iterations. The curves also saturate after a small number of steps, indicating that additional refinement alone cannot recover the gap created by missing reusable memory. 

\paragraph{Robustness to Irrelevant Memory.} We test whether our system blindly follows memory content when the inserted insights are not relevant to the image. This is a controlled stress test rather than the standard retrieval setting: instead of injecting the top-ranked memories, we deliberately insert the least-relevant insights from the memory bank. \cref{tab:memory-relevance} shows that these do not meaningfully hurt performance, remaining close to the no-memory setting across all metrics. In contrast, inserting the most-relevant memories substantially improves first-pass performance. This suggests that the gain comes from visually relevant correction insights, while irrelevant insights are largely ignored rather than blindly applied.
\\

\noindent The supplementary material provides further analyses demonstrating robustness to the processing order (Supp. \ref{app:robust}), the continued effectiveness of our method when using a different VLM (Supp. \ref{app:vlm_swap}), more qualitative examples (Supp. \ref{app:correction_examples}), audit of insights (Supp. \ref{app:audit}), efficiency and estimated costs (Supp. \ref{cost}), failure cases (Supp. \ref{failure_cases}), and limitations (Supp. \ref{limitations}).

\section{Conclusion}
We introduce CorrMem, a patch-indexed correction memory for guideline-consistent segmentation that converts successful correction episodes into reusable, visually grounded insights. Each insight pairs a directive natural-language correction insight with patch-level vision concept vectors extracted from the regions that caused the original error, allowing relevant lessons to help make the segmenting agent's first prediction on future images. Experiments on Waymo and Cityscapes show that this memory improves segmentation quality while reducing refinement passes. More broadly, our results show how vision agents can accumulate and reuse localized correction experience across inputs, shifting refinement from repeatedly rediscovering errors toward a persistent process that is proactive, more accurate, and more efficient.

\section*{LLM Usage Statement}
The authors wrote the first draft of this manuscript. Claude-opus-5.5 \cite{anthropic2026opus55} and GPT-6-astra \cite{openai2026gpt6astra} were used to polish the language and correct grammar. The research design and analyses were primarily the work of the authors. LLMs helped cross-check the hypotheses and results. During the ideation phase, LLMs also helped identify related work alongside the authors’ own literature search. No LLMs were used to create figures. Table layouts were refined with some assistance from LLMs. Claude Code was used to assist with coding experiments designed by the authors. Gemini-2.5-Flash \cite{comanici2025gemini} is part of the proposed system, as described in the main paper. Additional LLMs are used in the model-swap ablations and insight audit, as detailed in the supplementary material. The authors reviewed the resulting manuscript and take full responsibility for the final submission.

{
    \small
    \bibliographystyle{ieeenat_fullname}
    \bibliography{main}
}

%
\appendix
\clearpage

\twocolumn[
  {%
    \centering
    \bfseries\huge Supplementary Material\\[0.5em]\LARGE CorrMem: Visually Grounded Correction Memory for Guideline-Consistent Segmentation\\[1em]  
    \phantomsection
    \addcontentsline{toc}{section}{Supplementary Material}%
    \vspace{1em}                                 
  }%
]

\setcounter{section}{0}
\renewcommand{\thesection}{\Alph{section}}

\setcounter{table}{0}
\renewcommand{\thetable}{\Alph{table}}

\setcounter{figure}{0}
\renewcommand{\thefigure}{\Alph{figure}}

\section{Contents}
This supplementary material is organized as follows:
\begin{itemize}
    \item \ref{app:seg_guidelines}: Complete Segmentation Guidelines 
    \begin{itemize}[leftmargin=10pt]
        \item \ref{app:waymo_guide} Waymo Guidelines
        \item \ref{app:cityscapes_guide} Cityscapes Guidelines
    \end{itemize}
    \item \ref{app:cityscapes}: Cityscapes Results
    \item \ref{app:ablations}: More Ablations 
    \begin{itemize}[leftmargin=10pt]
        \item \ref{app:robust} Gains remain robust to the processing order 
        \item \ref{app:qual_patch_cls} Qualitative analysis of patch vs. global retrieval
        \item \ref{app:held_out} Frozen memory generalizes to unseen scenes
        \item \ref{app:vlm_swap} Gains persist even after swapping base VLMs
        \item \ref{app:cyclist} Memory gains extend to additional classes
        \item \ref{app:selective_vs_full} Selective retrieval performs better than using all insights
    \end{itemize}
    \item \ref{app:correction_examples}: Qualitative Examples of Correction Reuse
    \item \ref{app:audit}: Audit of Distilled Insights
    \item \ref{cost}: Estimated Costs
    \item \ref{failure_cases}: Examples of Failure Cases
    \item \ref{limitations}: Limitations and Future Work
\end{itemize}
\smallskip

\subsection{Complete Segmentation Guidelines}
\label{app:seg_guidelines}
Below are the segmentation guideline examples used in our experiments.
\subsubsection{Waymo.}
\label{app:waymo_guide}
In this study, we provide the system with the official Waymo guidelines as on their GitHub page. These detailed guidelines require several fine-grained decisions beyond basic category recognition.
\begin{itemize}
\small
    \item \texttt{All objects that can be recognized as a pedestrian and are at least partially visible are labeled.}
    \item \texttt{If it is not possible to tell by looking at the camera image whether an object is a pedestrian, the object is not labeled.}
    \item \texttt{People who are walking or riding kick scooters (including electric kick scooters), segways, skateboards, etc. are labeled as pedestrians.}
    \item \texttt{People inside other vehicles are not labeled, except for people standing on the top of cars/trucks or standing on flatbeds of trucks.}
    \item \texttt{A person riding a bicycle is not labeled as a pedestrian, but labeled as a cyclist instead.}
    \item \texttt{Mannequins, statues, billboards, posters, or reflections of people are not labeled.}
    \item \texttt{Single label (including the pedestrian and additional objects) is created if the pedestrian is holding a small child or carrying small items (smaller than 2m in size such as umbrella or small handbag or a sign) being held.}
    \item \texttt{Single label (including the pedestrian and additional objects) is created if the pedestrian is  riding a kick scooter (including electric kick scooter), a segway, a skateboard, etc.}
    \item \texttt{If the pedestrian is carrying an object larger than 2m, or pushing a bike or shopping cart, the label does not include the additional object.}
    \item \texttt{If the pedestrian is pushing a stroller with a child in it, separate bounding boxes are created for the pedestrian and the child. The stroller is not included in the child box.}
    \item \texttt{If pedestrians overlap each other, they are labeled as separate objects.}
\end{itemize}

\subsubsection{Cityscapes.}
\label{app:cityscapes_guide}
Similarly, for Cityscapes \textit{Person} class, we refer to their official guidelines:

\noindent \texttt{\small A human that satisfies the following criterion. Assume the human moved a distance of 1m and stopped again. If the human would walk, the label is person, otherwise not. Examples are people walking, standing or sitting on the ground, on a bench, on a chair. This class also includes toddlers, someone pushing a bicycle or standing next to it with both legs on the same side of the bicycle. This class includes anything that is carried by the person, e.g. backpack, but not items touching the ground, e.g. trolleys.}

\begin{figure*}[t]
  \centering
\includegraphics[width=\textwidth]{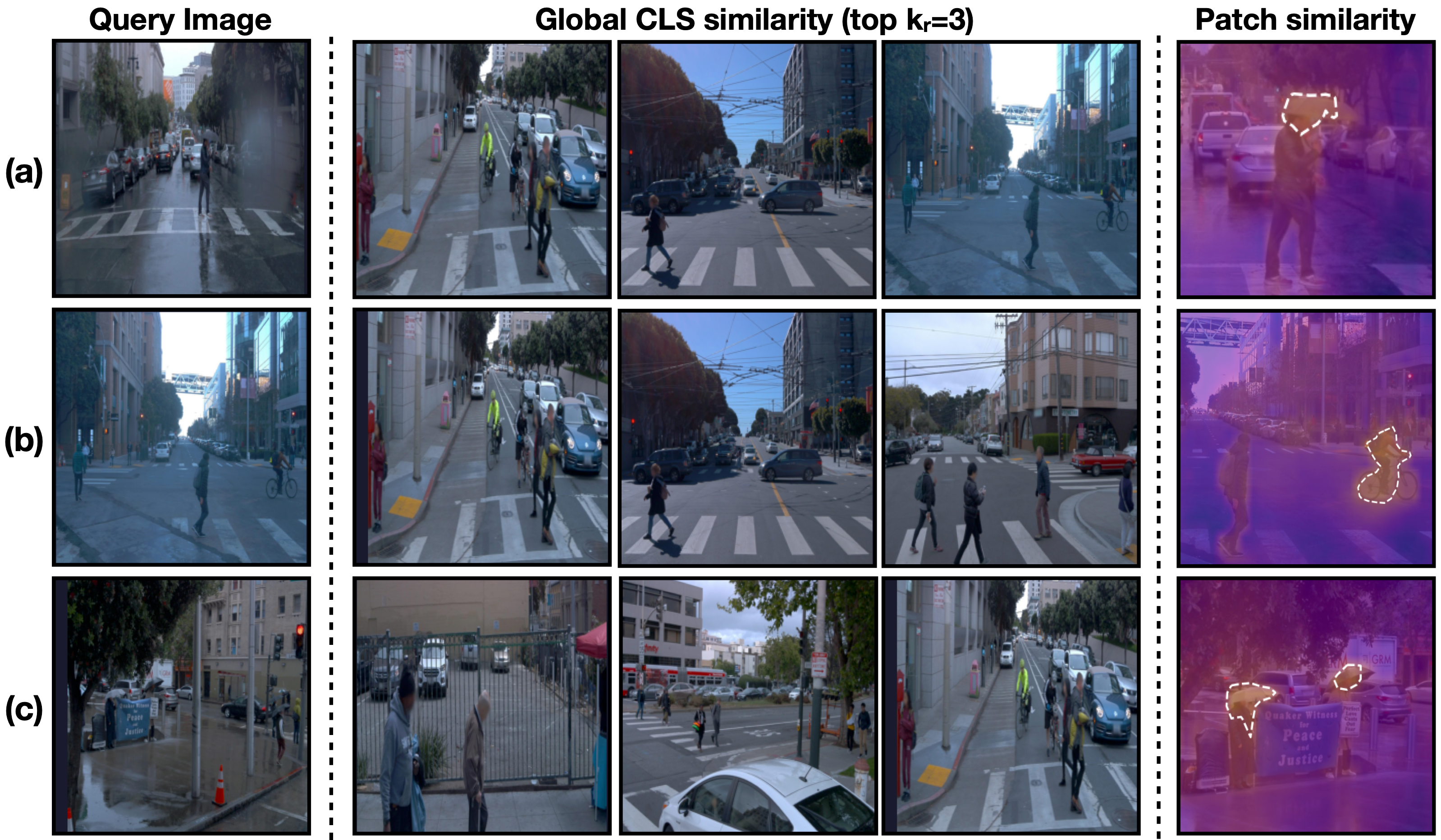}
  \caption{Qualitative comparison of patch vs global retrieval. Global CLS retrieval repeatedly returns similar street scenes, whereas patch similarity (zoomed in) localizes the guideline-relevant visual cues [(a),(c): umbrellas, (b): cyclist].}
  \label{fig:cls_vs_patch}
\end{figure*}

\begin{table}[ht]
\small
\renewcommand{\arraystretch}{1.2}
\centering
\caption{Effect of utilizing memories on the Cityscapes dataset. By using previous insights, our method improves the initial segmentation quality and further surpasses the refinement baseline after refinement on all metrics, while reducing the average refinement passes.}
\resizebox{\columnwidth}{!}{%
\begin{tabular}{@{}r|cc|cc@{}}
\toprule
\multirow{2}{*}{\diagbox[width=2.0cm,height=2.3\line,outerleftsep=1pt,outerrightsep=0pt,innerleftsep=0pt,innerrightsep=0pt]{Metrics}{Method}} &
\multicolumn{2}{c|}{Before Refinement} &
\multicolumn{2}{c}{After Refinement} \\
\cmidrule(l){2-3}\cmidrule(l){4-5}
& No memory & Ours & No memory & Ours \\
\midrule
gIoU $\uparrow$  & 49.20 & \textbf{51.91} & 51.14 & \textbf{55.71} \\
cIoU $\uparrow$ & 60.32 & \textbf{64.69} & 64.55 & \textbf{70.40} \\
mPr $\uparrow$  & 63.37 & \textbf{68.34} & 66.43 & \textbf{72.89} \\
mRec $\uparrow$ & 62.44 & \textbf{63.21} & 61.07 & \textbf{63.44} \\
mDice $\uparrow$ & 60.33 & \textbf{63.24} & 61.72 & \textbf{66.35} \\
\midrule
\rowcolor[HTML]{EFEFEF}
\multicolumn{1}{r|}{Avg \#iters $\downarrow$} &
- & - &
2.33 & \textbf{1.23} \footnotesize{\textit{($\downarrow1.9\times$)}}\\
\bottomrule
\end{tabular}%
}
\label{tab:num_iter_cityscapes}
\end{table}

\subsection{Cityscapes Results}
\label{app:cityscapes}
\cref{tab:num_iter_cityscapes} extends the memory-effect analysis from \cref{tab:num_iter} in the main paper to the Cityscapes validation set. We compare the same segmentation pipeline with and without reusable insights, both before and after refinement. Memory improves gIoU, cIoU, precision, recall, and Dice at both stages while reducing the average number of iterations from 2.33 to 1.23. On this dataset too, the first-pass gIoU improvement is statistically reliable with +2.7 gIoU (paired-bootstrap 95\% CI [+1.2, +4.3]; Wilcoxon signed-rank $p<0.001$). This shows that the central benefit of correction reuse extends beyond the Waymo benchmark. Absolute metric numbers are lower on Cityscapes because many person instances are very small and distant, making them harder to detect.


\subsection{More Ablations}
\label{app:ablations}
We include some more ablations and analyses to further support our claims.

\subsubsection{Memory gains remain robust to the processing order.} 
\label{app:robust} 
Since our memory is built online, later samples benefit from insights learned from earlier correction episodes. This raises a natural question: whether the method depends strongly on the order in which samples are processed. \cref{tab:split-comparison-before-after} evaluates this effect by comparing the original dataset order with a randomly shuffled order. The final results are very similar across both settings and require the same average number of refinement iterations. This suggests that the memory benefits are not tied to a particular sample order for a whole dataset; instead, the stored insights capture recurring correction patterns that remain useful across different processing sequences.

\begin{table}[t]
\centering
\caption{Effect of processing order. Similar performance under the original and randomly shuffled orders indicates that memory benefits are not order-dependent.}
\footnotesize
\renewcommand{\arraystretch}{1.0}
\setlength{\tabcolsep}{4pt}
\begin{tabularx}{\columnwidth}{@{}r|>{\centering\arraybackslash}X>{\centering\arraybackslash}X|>{\centering\arraybackslash}X>{\centering\arraybackslash}X@{}}
\toprule
\multirow{2}{*}{\diagbox[width=1.8cm,height=2.5\line,outerleftsep=1pt,outerrightsep=0pt,innerleftsep=1pt,innerrightsep=0pt]{Metrics}{Method}} &
\multicolumn{2}{c|}{Before Refinement} &
\multicolumn{2}{c}{After Refinement} \\
\cmidrule(l){2-3}\cmidrule(l){4-5}
& Original & Random & Original & Random \\
\midrule
gIoU  & 78.85 & 77.24 & 83.51 & 83.24 \\
cIoU  & 82.88 & 82.09 & 87.68 & 86.43 \\
mPr   & 90.31 & 88.31 & 94.44 & 94.26 \\
mRec  & 86.58 & 85.97 & 87.55 & 86.71 \\
mDice & 88.40 & 87.12 & 90.21 & 90.33 \\
\midrule
\multicolumn{1}{r|}{Avg \#iters} &
- & - & 1.01 & 1.01 \\
\bottomrule
\end{tabularx}
\label{tab:split-comparison-before-after}
\end{table}

\subsubsection{Qualitative analysis of patch-level vs. global retrieval.} 
\label{app:qual_patch_cls}
\cref{tab:cls_patch_ablation} in the main paper quantitatively demonstrates the advantage of patch-based retrieval over global CLS retrieval. \cref{fig:cls_vs_patch} provides a qualitative explanation. Queries (a) and (c) contain umbrella-related cues, while (b) contains a cyclist. Global CLS similarity is dominated by overall street appearance, causing the same memory image to appear among the top three results for all queries despite their different local triggers. Patch similarity instead concentrates on the relevant regions, outlined in white, enabling more cue-specific retrieval.

\subsubsection{Frozen memory generalizes to held-out unseen scenes.}
\label{app:held_out}
To assess whether accumulated correction experience transfers to unseen scenes under the same annotation guidelines, we evaluate on 50 held-out Waymo images randomly selected using the same selection criteria as GuideSeg \cite{vats2026guideline}. These images come from scenes disjoint from the guideline-consistent set used to construct memory. We reuse the 22-insight bank accumulated during the original guideline-consistent experiment and freeze it throughout evaluation: test images neither contribute new insights nor update existing entries. All other configurations and parameters remain unchanged from the original experiments.

On this held-out test set, the no-memory pipeline scores 71.16 first-pass gIoU, but our stateful pipeline improves performance to 78.82 gIoU. Our method shows similar improvements in other metrics too (+3.74 mPr, +4.58 mRec, +7.19 mDice). These results support the transfer of previously acquired correction knowledge to unseen scenes under the same guidelines, without further memory accumulation during evaluation.

\begin{table}[t]
\centering
\caption{Effect of swapping the base VLM. Patch-based memory improves overall first-pass and final segmentation (after refinement) performance across gemini-2.5-flash, gemini-3-flash-preview, and gpt-6.1-terra while reducing refinement iterations, indicating that its benefits are not limited to the primary VLM configuration.}
\renewcommand{\arraystretch}{1.2}
\setlength{\tabcolsep}{3pt}
\resizebox{\columnwidth}{!}{%
\begin{tabular}{c|c|c|ccccc|c}
\toprule
Model & Ref. & Mem. & gIoU$\uparrow$ & cIoU$\uparrow$ & mPr$\uparrow$ & mRec$\uparrow$ & mDice$\uparrow$ & \#iter$\downarrow$\\
\midrule
\multirow{4}{*}{\shortstack{gemini-2.5-\\flash}}
 & \multirow{2}{*}{Before}  & \no  & 73.87 & 78.40 & 88.36 & 80.81 & 84.75 & -\\
 &                         & \yes & \textbf{78.85} & \textbf{82.88} & \textbf{90.31} & \textbf{86.58} & \textbf{88.40} & -\\
 \cmidrule(l){2-9}
 & \multirow{2}{*}{After} & \no  & 80.57 & 86.70 & 91.06 & 84.78 & 87.20 & 2.66\\
 &                        & \yes & \textbf{83.51} & \textbf{87.68} & \textbf{94.44} & \textbf{87.55} & \textbf{90.21} & \textbf{1.01}\\
\midrule
\multirow{4}{*}{\shortstack{gemini-3-\\flash}}
 & \multirow{2}{*}{Before}  & \no  & 75.97 & 78.70 & 91.12 & 79.42 & 84.00 & -\\
 &                         & \yes & \textbf{82.65} & \textbf{85.45} & \textbf{93.92} & \textbf{87.41} & \textbf{89.56} & -\\
 \cmidrule(l){2-9}
 & \multirow{2}{*}{After} & \no  & 79.36 & 85.85 & 88.66 & 83.64 & 85.47 & 2.86\\
 &                        & \yes & \textbf{84.22} & \textbf{89.34} & \textbf{93.56} & \textbf{87.74} & \textbf{90.16} & \textbf{1.44}\\

 \midrule
\multirow{4}{*}{\shortstack{gpt-6.1-\\terra}}
 & \multirow{2}{*}{Before}  & \no  & 83.94 & 87.92 & 94.51 & 88.35 & 90.57 & -\\
 &                         & \yes & \textbf{85.26} & \textbf{89.40} & 94.49 & \textbf{89.36} & \textbf{91.28} & -\\
 \cmidrule(l){2-9}
 & \multirow{2}{*}{After} & \no  & 85.40 & 89.91 & 93.66 & 90.63 & 91.37 & 1.23\\
 &                        & \yes & \textbf{85.83} & \textbf{90.62} & \textbf{94.40} & 90.35 & \textbf{91.61} & \textbf{0.45}\\
 
\bottomrule
\end{tabular}%
}
\label{tab:gemini3}
\end{table}
 \begin{figure}[ht]
  \centering
\includegraphics[width=\columnwidth]{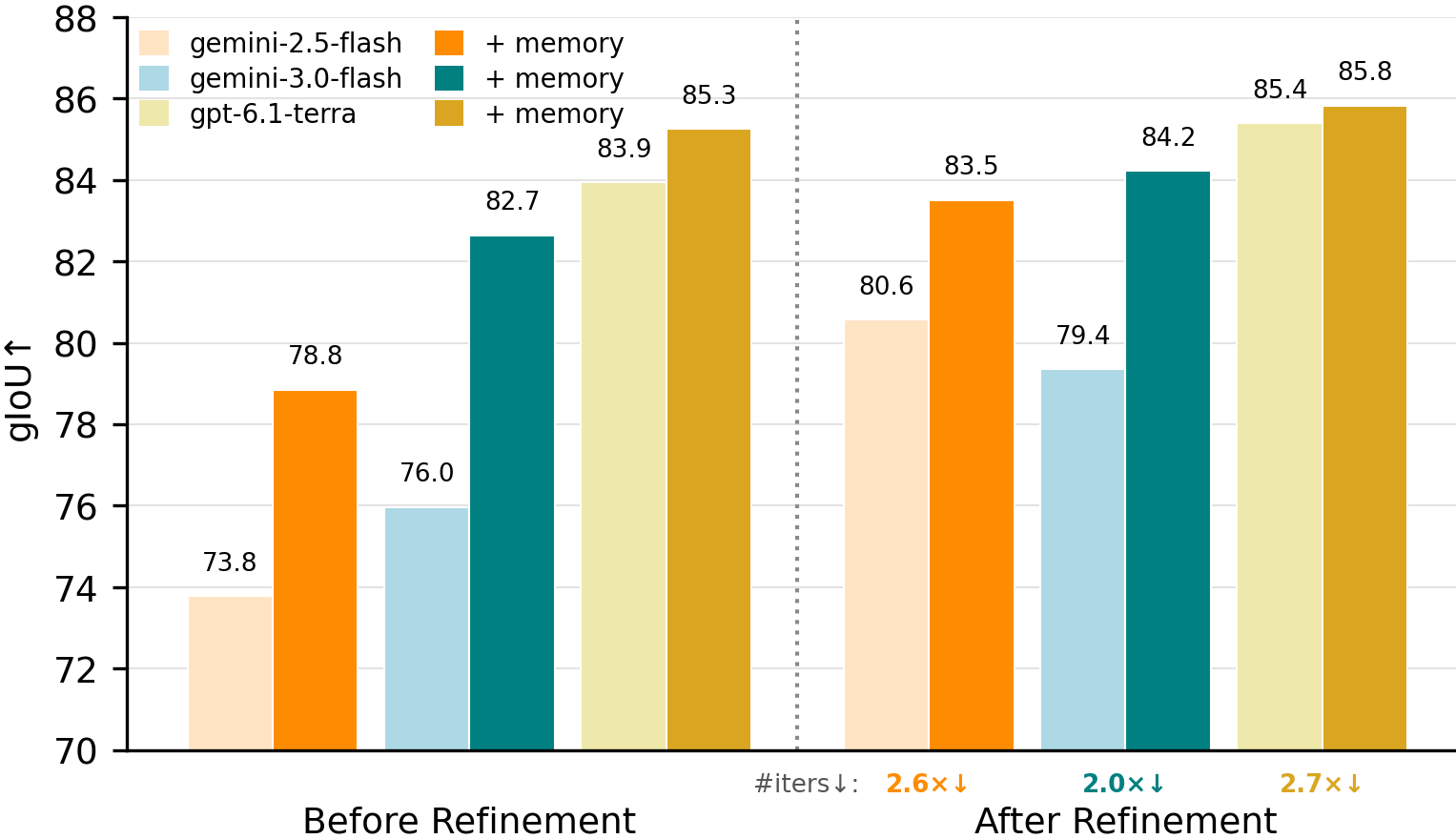}
  \caption{Effect of swapping the base VLM on gIoU. Our system improves both first-pass and final performance for all three models tested while reducing refinement iterations, showing that the gains are not restricted to the primary VLM settings.}
  \label{fig:memory_robustness}
\end{figure}

\subsubsection{Gains persist even after swapping base VLMs.}
\label{app:vlm_swap}
We use gemini-2.5-flash as the primary VLM to enable an apples-to-apples comparison with the stateless baseline. To examine whether the memory gains are specific to this model, we replace it with gemini-3-flash-preview \cite{google2025gemini3flash} while keeping the datasets, guidelines, and remaining pipeline unchanged. Additionally, we evaluate on gpt-6.1-terra \cite{openai2026gpt61terra} to test whether these gains extend beyond the Gemini family. As shown in \cref{tab:gemini3} and \cref{fig:memory_robustness}, incorporating our patch-based memory generally produces better first-pass and final segmentation performance and efficiency with both newer VLMs, including the stronger gpt-6.1-terra. These results indicate that the benefits of reusable correction insights are not limited to the primary VLM configuration.

\begin{figure*}[h!]
  \centering
\includegraphics[width=0.95\textwidth]{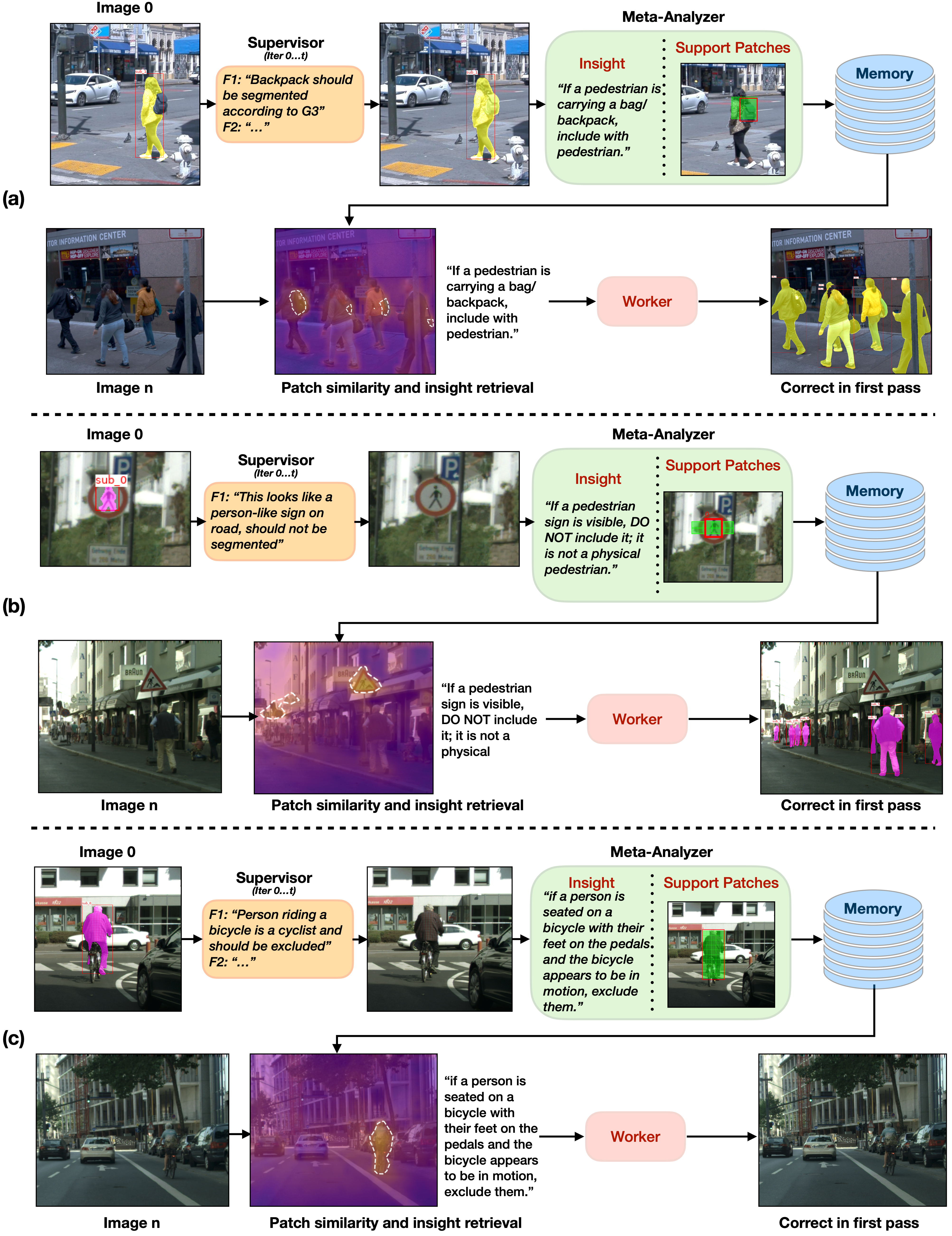}
  \caption{Examples of writing and retrieving visually grounded correction insights for guideline-consistent segmentation.}
  \label{fig:qual_pipeline_examples}
\end{figure*}

\subsubsection{Memory gains extend to additional classes.} 
\label{app:cyclist}
To test whether the benefit of correction memory extends beyond pedestrians, we evaluate CorrMem with no-memory baseline \cite{vats2026guideline} on the \textit{Cyclist} class using the same configuration and settings. We randomly select a subset of 30 Waymo images for this evaluation using the same selection criteria as GuideSeg \cite{vats2026guideline}, and the cyclist labeling guidelines are provided below. Our method improves first-pass gIoU from 29.16 to 40.56$\pm$1.03 and Dice from 39.31 to 50.24$\pm$1.23. These improvements are statistically significant (Wilcoxon signed-rank $p \leq 0.023$), and their paired 95\% bootstrap confidence intervals exclude zero. The final segmentation performance after refinement also improves by +8.82 gIoU and +10.95 mDice, while requiring fewer refinement iterations per image (1.84$\pm$0.17 versus 2.72).
The larger gains compared with the pedestrian class may reflect recurring cyclist-specific decisions, particularly bicycle inclusion and mounting or dismounting status, which a small set of correction insights can capture and reuse across scenes. The lower memory-free starting performance on the cyclist subset also leaves greater room for improvement. Nevertheless, these results provide additional evidence that the correction memory could prove helpful on other classes as well. 

The Cyclist guidelines are same as found on the official Waymo labeling page:
\begin{itemize}
\small
    \item \texttt{Bicycles that are parked or do not have a rider are not labeled.}
    \item \texttt{When a person is getting onto a bicycle, they are labeled as person until they are about to get onto the bicycle, and labeled as cyclist after the rider gets into the riding position.}
    \item \texttt{When a person is getting off of a bicycle, they are labeled as cyclist while the rider is in the riding position. If the person is walking with the bicycle, they are not labeled as cyclist.}
    \item \texttt{Labels are created for persons riding bicycles, unicycles, tricycles, recumbent bicycles, and large, multi-seat cycles.}
    \item \texttt{Label children riding bicycles, tricycles, or wheeled toys as cyclists.}
    \item \texttt{The label includes both the rider and the bicycle.}
    \item \texttt{Objects attached to the rear of an active cyclist, such as seats and small trailers, are included in the label.}
\end{itemize}

\subsubsection{Selective retrieval performs better than using all insights.}
\label{app:selective_vs_full}
We test whether selectively retrieving relevant insights is necessary, or whether the worker benefits equally from receiving every insight stored so far. In the same online setting as the rest of our experiments, instead of injecting the top-$k$ retrieved insights, we inject all insights accumulated before the current sample, with the rest of the settings unchanged.  

This setting provides all applicable insights already in memory, alongside potentially irrelevant or redundant ones. The worker thus has access to the relevant corrections, but must determine which to apply from a bloated context mixing relevant and irrelevant insights. As expected, injecting all insights improves first-pass performance over the no-memory pipeline (73.87 to
77.58 gIoU, 78.40 to 79.88 cIoU, 84.75 to 87.32 mDice). But this still falls short of our method, with at most $k_r=3$ insights, performing better with 78.85 gIoU, 82.88 cIoU, 88.40 mDice. Together with \cref{tab:memory-relevance} in the main paper, where inserting the least-relevant insights alone performs close to the no-memory baseline, this suggests that additional, less relevant context can weaken the benefit of useful corrections. Selecting the top-$k$ patch-indexed insights avoids this interference while keeping the worker's context bounded. We use $k_r=3$ throughout these experiments as a compact operating point, but the retrieval budget remains configurable, allowing users to balance coverage of potentially useful corrections against context size.


\subsection{Qualitative Examples of Correction Reuse}
\label{app:correction_examples}
\cref{fig:qual_pipeline_examples} illustrates three examples of cross-image correction reuse. In each case, the Supervisor first identifies a guideline violation, after which the Meta-Analyzer converts the successful correction into a reusable insight and anchors it to the corresponding support patches. When a similar visual cue appears in a later image, patch-level retrieval recalls the relevant insight before segmentation, allowing the Worker to produce the correct mask in its first pass. The examples cover: (a) including a carried backpack, (b) excluding a person depicted on a traffic sign, and (c) excluding a cyclist from the pedestrian mask.

\begin{figure*}[t!]
  \centering
\includegraphics[width=\textwidth]{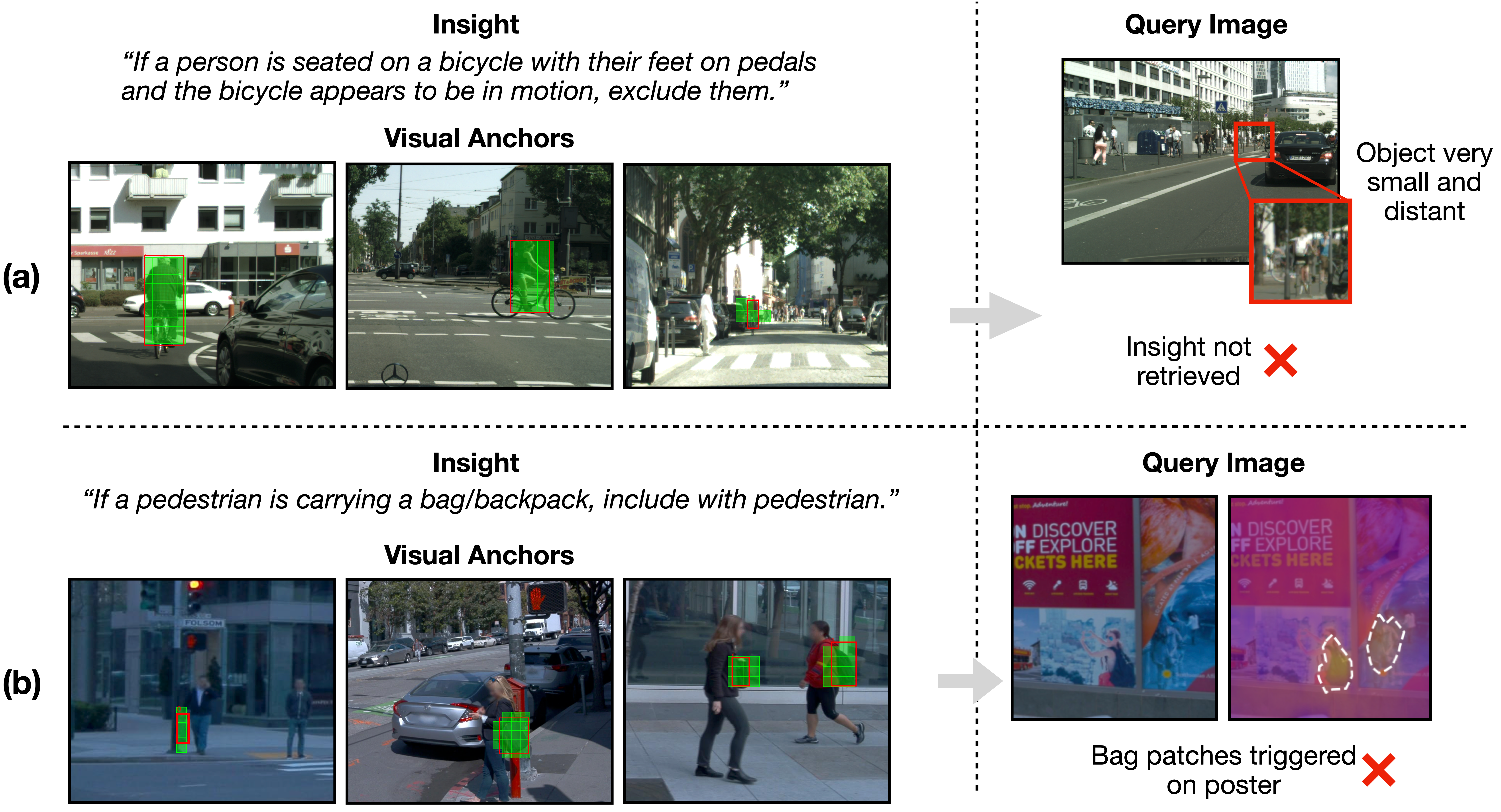}
  \caption{Some examples of failure cases. (a) A small and distant cyclist does not sufficiently match the stored cyclist anchors, so the relevant insight is missed. (b) Person and bag depictions on a poster resemble the stored bag anchors, causing a false retrieval.}
  \label{fig:failure}
\end{figure*}

\subsection{Audit of Distilled Insights}
\label{app:audit}
We audit all insights distilled from qualifying correction episodes on Waymo ($n_{\mathrm{ins}}=42$) and Cityscapes ($n_{\mathrm{ins}}=104$). For each insight, auditors inspect the initial and final segmentations, the supervisor feedback, and the annotation guidelines. They assess two questions: whether the supervisor feedback is correct given the image and guidelines, 
  \begin{table}[h!]
  \centering
  \small
  \caption{Quality of insight distillation, audited by human and LLM as a judge in \%. All three evaluators rate most insights as correct and largely agree with each other.}
  \label{tab:insight_audit}
  \begin{tabular}{l|l|ccc}
  \toprule
  Dataset & Verdict & Human & LLM1 & LLM2 \\
  \midrule
  \multirow{3}{*}{Waymo}  & Correct  & 90.5 & 88.1 & 95.2 \\
                                & Partly correct &  00.0  &  02.4  &  00.0  \\
                             & Incorrect      &  09.5  &  09.5  &  04.8  \\
  \midrule
  \multirow{3}{*}{Cityscapes} & Correct & 88.5 & 84.6 & 89.4 \\
                                          & Partly correct &  03.8  &  04.8  &  02.9  \\
                                          & Incorrect      &  07.7  & 10.6 &  07.7  \\
  \bottomrule
  \end{tabular}
  \end{table} and whether the insight faithfully reflects that feedback while following the guidelines. An insight is \textit{correct} if both judgments are positive, \textit{partly correct} if either is partially positive and neither is negative, and \textit{incorrect} if either is negative.

\noindent One author performs the blind human audit independently. We also use two LLM judges: LLM1 (claude-opus-5-5 \cite{anthropic2026opus55}) and LLM2 (gpt-6-astra \cite{openai2026gpt6astra}), both with \texttt{xhigh} thinking. Note that these models are different from the Gemini class model we are using for producing insights. The human auditor rates approximately 90\% of insights as correct on both datasets (90.5\% on Waymo and 88.5\% on Cityscapes), with fewer than one in ten rated incorrect. The LLM judges produce similar results (\cref{tab:insight_audit}). Pairwise agreement among the three evaluators ranges from 88-93\% on Waymo and 84-86\% on Cityscapes. All three evaluators judge that the insights faithfully follow their cited feedback in at least 96\% of cases, so the few incorrect ones reflect the feedback the meta-analyzer receives rather than the distillation itself. We retain the baseline supervisor unchanged to isolate the contribution of memory. The audited insights come from the same runs as our main results, so the reported gains already account for the presence of incorrect insights. \cref{tab:memory-relevance} further demonstrates robustness to low-relevance insights. Since the meta-analyzer is not tied to a particular supervisor, improving feedback accuracy could further improve insight quality, which we leave to future work.

\subsection{Estimated Costs}
\label{cost}
Beyond improving segmentation quality, CorrMem reduces API usage by preventing unnecessary refinement. We use stable gemini-2.5-flash, priced at USD\$0.30 per million input tokens and USD\$2.50 per million output tokens. At approximately 2,000 input and 200 output tokens, each Worker or Supervisor call costs about \$0.0011. A complete Worker-Supervisor refinement iteration requires three API calls, whereas memory insertion and retrieval are performed locally using FAISS and incur no additional API cost. The complete serialized memory banks, including insights, visual anchors, and associated metadata, occupy only 869 KB for Waymo and 1.55 MB for Cityscapes. In addition, retrieved insights reduce the average number of refinement iterations from 2.66 to 1.01 on Waymo. Writing an insight requires one batched meta-analyzer call, which occurs only for the 28.6\% of samples that yield a qualifying correction episode, adding less than a tenth of an iteration's cost per sample. The memory-free system therefore averages $7.98$ refinement calls (\$0.00878) per sample, while CorrMem averages $3.32$ calls (\$0.00365), including memory writes. As a result, the API refinement cost reduces by 2.4$\times$, from \$8.78 to \$3.65 per 1,000 samples.

\subsection{Examples of Failure Cases}
\label{failure_cases}
\cref{fig:failure} shows two representative limitations of our method. In (a), the memory contains an applicable cyclist insight, but the cyclist in the query is too small and distant to preserve sufficient visual similarity with its anchors. In (b), local patches depicting people and bags on a poster resemble the stored bag anchors, even though they do not correspond to physical pedestrians in the scene. These cases show that local matching can miss cues under substantial scale changes or retrieve visually similar cues without sufficient scene context. Multi-scale representations and semantic verification of retrieved cues could help address these limitations.


\subsection{Limitations and Future Work}
\label{limitations}
As an online memory system, CorrMem can reuse a correction only after encountering a related case; until then, it falls back to the underlying Worker-Supervisor refinement loop. Every online run begins with an empty memory bank, so the observed gains already include this cold-start period and do not depend on preloaded insights. Our experiments focus on dataset-scale memory, while longer deployments may accumulate insights that become redundant or stale as visual conditions and labeling policies evolve, a case that text-level deduplication does not address. 

Because insights are stored as explicit natural-language records rather than changing model weights, the memory bank remains directly inspectable and auditable. This creates a practical future path toward insight-level confidence tracking, manual auditing, and selective consolidation or retirement of stale insights without retraining the underlying models.

\end{document}